\documentclass[journal,twoside,web]{ieeecolor}
\usepackage{tmi}
\usepackage{cite}
\usepackage{hyperref}
\hypersetup{colorlinks,linkcolor=black,citecolor=black,urlcolor=black} 
\usepackage{amsmath,amssymb,amsfonts}
\usepackage{algorithmic}
\usepackage{graphicx}
\usepackage{textcomp}
\usepackage{subfigure}
\usepackage{tabu}
\usepackage{multirow}
\usepackage{multicol}
\usepackage{float}
\usepackage{makecell}
\usepackage{booktabs}
\usepackage{pifont}
\usepackage{CJKutf8}

\def\BibTeX{{\rm B\kern-.05em{\sc i\kern-.025em b}\kern-.08em
    T\kern-.1667em\lower.7ex\hbox{E}\kern-.125emX}}
\begin{document}

\title{UniChest: Conquer-and-Divide Pre-training for Multi-Source Chest X-Ray Classification}

\author{Tianjie Dai, Ruipeng Zhang, Feng Hong, Jiangchao Yao, Ya Zhang, Yanfeng Wang
\thanks{This work is supported by National Key R\&D Program of China (No. 2022ZD0160702),  STCSM (No. 22511106101, No. 22511105700, No. 21DZ1100100), 111 plan (No. BP0719010), NSFC (No. 62306178).}
\thanks{T. Dai and Feng Hong are with the Cooperative Medianet Innovation Center, Shanghai Jiao Tong University, Shanghai, China.}
\thanks{R. Zhang, J. Yao, Ya Zhang and Y. Wang are with the Cooperative Medianet Innovation Center, Shanghai Jiao Tong University and with the Shanghai AI Laboratory, Shanghai, China. (Corresponding authors: Jiangchao Yao and Yanfeng Wang. Corresponding e-mails: Sunarker@sjtu.edu.cn, wangyanfeng622@sjtu.edu.cn)}
\thanks{© 2024 IEEE.  Personal use of this material is permitted.  Permission from IEEE must be obtained for all other uses, in any current or future media, including reprinting/republishing this material for advertising or promotional purposes, creating new collective works, for resale or redistribution to servers or lists, or reuse of any copyrighted component of this work in other works.}
}

\maketitle

\begin{abstract}
Vision-Language Pre-training~(VLP) that utilizes the multi-modal information to promote the training efficiency and effectiveness, has achieved great success in vision recognition of natural domains and shown promise in medical imaging diagnosis for the Chest X-Rays~(CXRs). 
However, current works mainly pay attention to the exploration on single dataset of CXRs, which locks the potential of this powerful paradigm on larger hybrid of multi-source CXRs datasets.
We identify that although blending samples from the diverse sources offers the advantages to improve the model generalization, it is still challenging to maintain the consistent superiority for the task of each source due to the existing heterogeneity among sources. 
To handle this dilemma, we design a Conquer-and-Divide pre-training framework, termed as UniChest, aiming to make full use of the collaboration benefit of multiple sources of CXRs while reducing the negative influence of the source heterogeneity. Specially, the \textit{``Conquer"} stage in UniChest encourages the model to sufficiently capture multi-source common patterns, and the \textit{``Divide"} stage helps squeeze personalized patterns into different small experts (query networks). We conduct thorough experiments on many benchmarks, {\em e.g.}, ChestX-ray14, CheXpert, Vindr-CXR, Shenzhen, Open-I and SIIM-ACR Pneumothorax, verifying the effectiveness of UniChest over a range of baselines, and release our codes and pre-training models at \url{https://github.com/Elfenreigen/UniChest}.
\end{abstract}


\section{Introduction}
\label{sec:introduction}

\IEEEPARstart{C}{hest} X-Ray~(CXR) in screening chest diseases is essential to detect and control their fatal infectious impact on human lives in broad countries~\cite{kim2020role}. To reduce the labor costs, deep learning techniques have become prevalent for machine-assisted CXR diagnosis, driving medical imaging recognition into the new era~\cite{shin2023impact}. Specially, with the rapid development of pre-training models, extensive studies have been conducted and shown promise in a wide range of tasks and domains~\cite{wang2023internimage}, drawing increasing attention in medical community.

Recently, Vision-Language Pre-training techniques has significantly improved the performance of machine-aided CXR disease diagnosis~\cite{boecking2022making, tiu2022expert,  wu2023medklip, zhang2023knowledge}.
Some studies even have shown the potential of surpassing the experienced radiologists in diagnosing some chest diseases~\cite{zhang2023knowledge}.
Besides, in combination with medical domain-specific knowledge, these pretrained models exhibit more reasonable explanations in the lesion grounding~\cite{wu2023medklip, zhang2023knowledge}.
Nevertheless, it is worth noting that these works for CXR VLP only consider pre-training on a single-source dataset \textit{e.g.,} MIMIC-CXR of about 300K samples~\cite{johnson2019mimic}.
Recalling the practice of GPT~\cite{dale2021gpt} or CLIP~\cite{radford2021learning} that utilizes billions of multi-source samples, single-source VLP inherently induces drawbacks in the disease coverage and representativeness~\cite{johnson2019mimic}, especially under real-world medical applications. 

\begin{figure*}
\centering
  \begin{minipage}{0.45\textwidth}
    \centering
    \includegraphics[width=\textwidth]{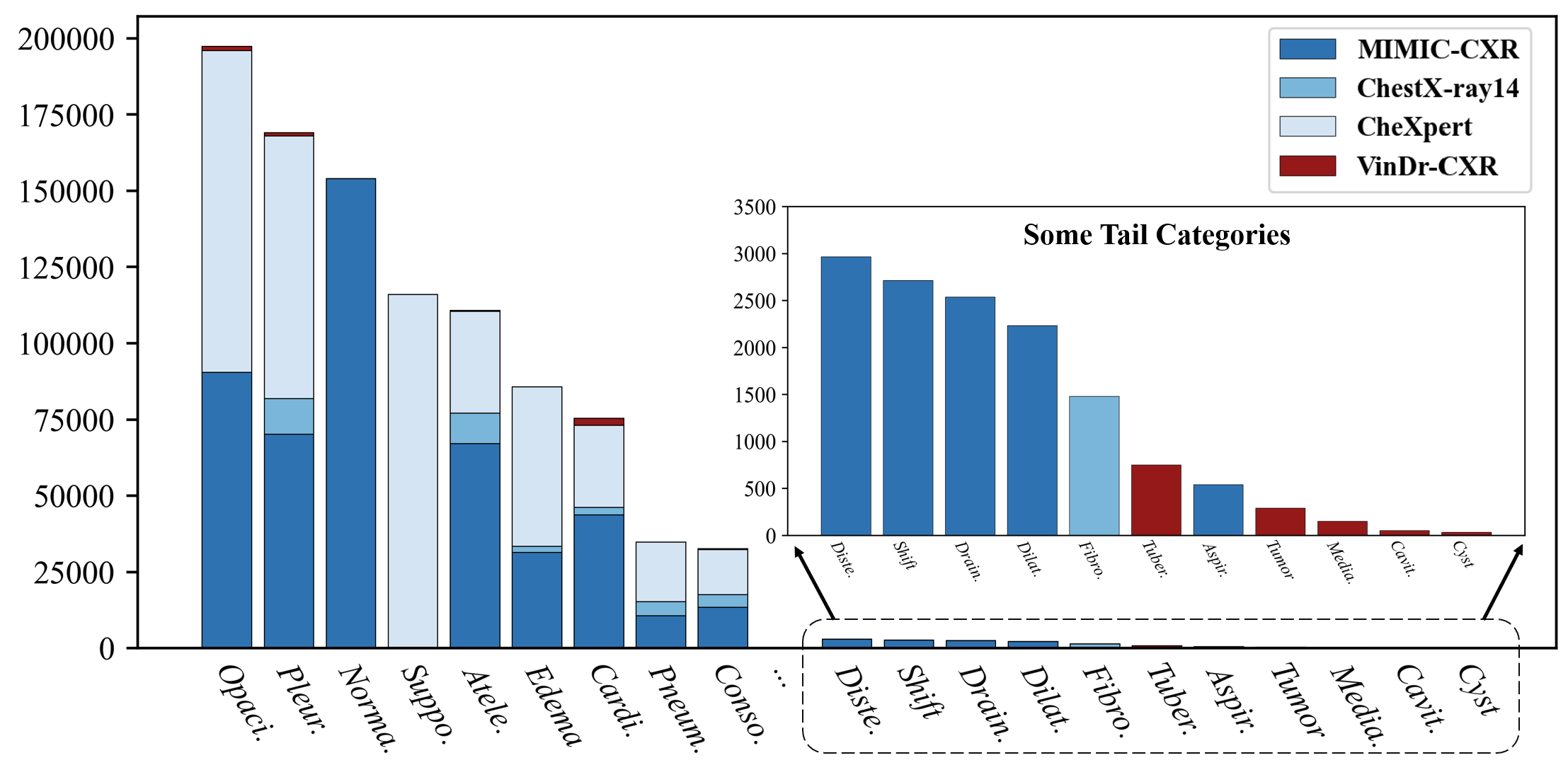}
  \end{minipage}
  \begin{minipage}{0.45\textwidth}
    \centering
    \includegraphics[width=0.95\textwidth]{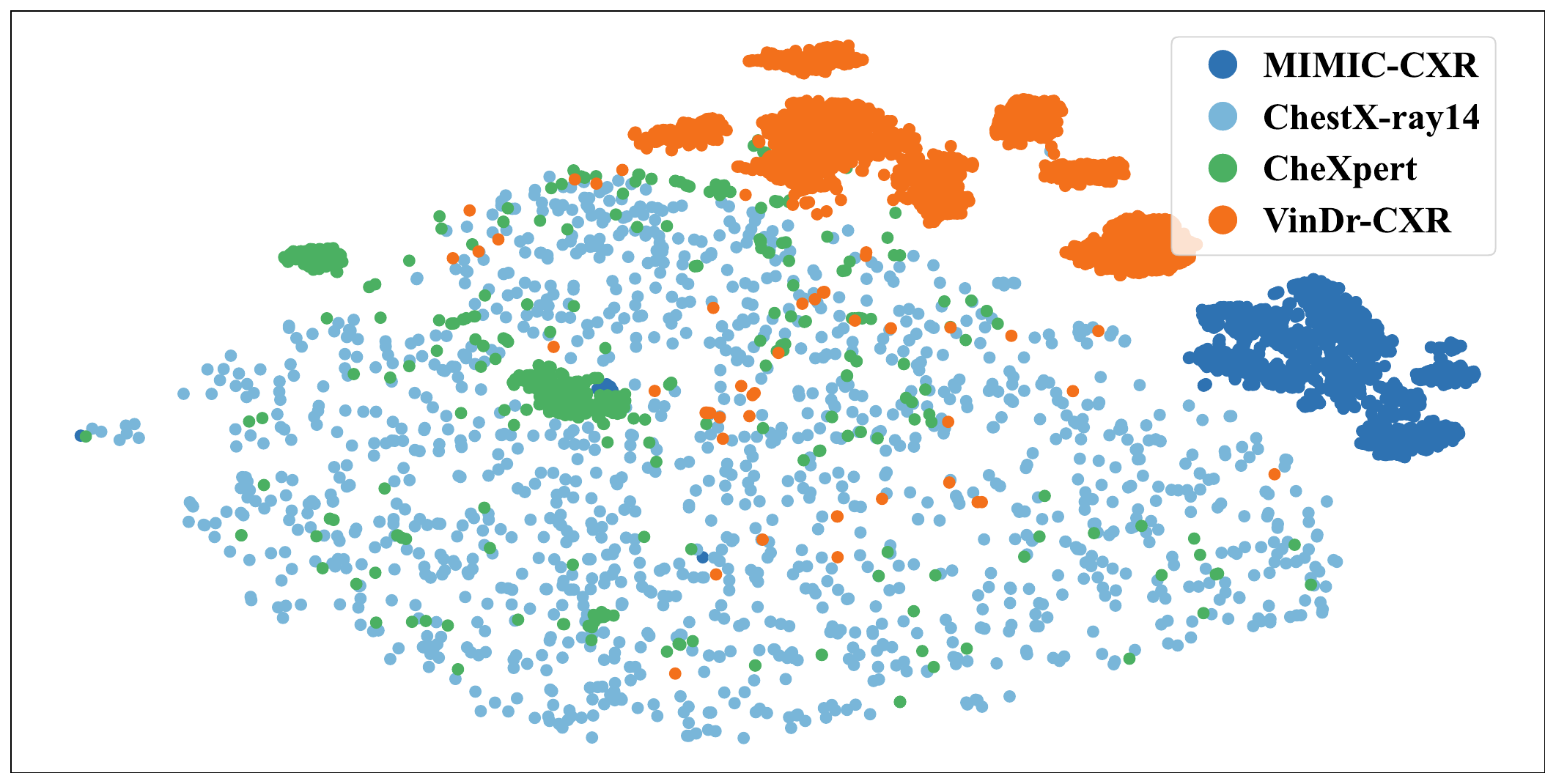}
  \end{minipage}

\caption{Left: The class distribution of the multi-source dataset composed of MIMIC-CXR, ChestX-ray14, CheXpert and VinDr-CXR training sets. The initial five characters of each label serve as the abbreviated x-axis annotation. Right: T-SNE visualization \textit{w.r.t.} the visual representations of medical images randomly selected from MIMIC-CXR, ChestX-ray14, CheXpert and VinDr-CXR, which characterizes the source heterogeneity.}
\label{motivation}
\end{figure*}

Motivated by the above limitation, we explore building a more powerful pre-training framework by leveraging multi-source CXR data. Generally, the label space union of multiple sources can help expand the coverage of the disease categories, particularly for rare diseases. Besides, samples from different sources with diverse radiation equipment, collection standards and population distributions, may complement each other~\cite{yang2022multi}, which helps enhance the generalization ability of pretrained models.
Unfortunately, we argue that it is still very challenging to effectively utilize multi-source CXRs, as the source heterogeneity (shown in Fig.~\ref{motivation}) also exacerbates the complexity of the CXR disease data, which could impair the holistic improvement for all source tasks during pre-training.

To address this dilemma, we propose a Conquer-and-Divide pre-training framework, termed as UniChest, which maintains the merits of multi-source data collaboration and simultaneously weakens the negative effect of the source heterogeneity by a proper training isolation. Specifically, at the ``Conquer" stage of UniChest, we promote the capture of multi-source common patterns at first in parts of the model, which focuses on enhancing the feature extraction abilities. 
Then, at the ``Divide" stage of UniChest, we introduce a mixture of experts warmed up from previous stage, together with the guidance of source contrastive learning, to squeeze the source-specific patterns into different experts, which reduces the interference among sources during pre-training. With this framework, we can achieve better efficiency and effectiveness for pre-training on the large-scale multi-source CXR data.
In a nutshell, the contribution of this work can be summarized as follows:
\begin{itemize}
\item We explore pre-training on large-scale multi-source CXR data and propose a Conquer-and-Divide pre-training framework to overcome the dilemma induced by the source heterogeneity in scaling up the data.
\item We design a mixture of deep query networks together with a source contrastive learning loss to squeeze source-specific patterns into separate components, promoting a harmonious multi-source collaboration for pre-training.
\item We conduct thorough experiments to show the promise of UniChest on multiple datasets, {\em i.e.}, ChestX-ray14~\cite{wang2017chestx}, CheXpert~\cite{irvin2019chexpert}, Vindr-CXR~\cite{nguyen2022vindr}, Shenzhen~\cite{jaeger2014two}, Open-I~\cite{demner2016preparing} and SIIM-ACR Pneumothorax~\cite{kaggle-siim}, achieving new state-of-the-art performance for diverse CXR diagnosis.
\end{itemize}

\section{Related Work}

\subsection{Deep Learning for Chest X-Ray Disease Diagnosis}
Considering the labor and repeatability of human experts in CXR diagnosis, it is possible to find the computer-aided solutions powered by deep learning~\cite{van2001computer, hosny2018artificial, chan2020computer, kieu2020survey}. Thereby, extensive CNN-based methods for CXRs have been explored in recent years~\cite{meedeniya2022chest}. For instance, Khoiriyah~\cite{khoiriyah2020convolutional} built a network comprising of three convolutional layers and three connected layers, showing remarkable performance in automatic pneumonia detection. The enhancement by transfer learning further reduced the training cost and improved the generalization performance~\cite{yang2020transfer, lee2020darwin}. Specifically, several studies utilized models pre-trained on the natural domain \textit{e.g.,} ImageNet~\cite{deng2009imagenet} as initialization and finetuned the last layer~\cite{choudhuri2021multi}. In addition, post-hoc techniques can be also very beneficial to enhance the stability and accuracy of the CXR diagnosis. For example, some explorations~\cite{hashmi2020efficient,chouhan2020novel} focus on the model ensemble, \textit{i.e.}, directly summarizing the multiple outputs of a series of models as the final prediction, achieving the remarkable performance.

\subsection{Vision-Language Pre-training in Medical Domain}
Recently, Vision-Language Pre-training~(VLP) models have achieved impressive success in natural domain~\cite{bianchi2021contrastive, jia2021scaling, li2021align}, which then drives many extensions in the medical area and improves the ability of machine-aided medical applications. ConVIRT~\cite{zhang2020contrastive} made the first attempt to integrate VLP into medical models, which follows the two-stream paradigm with the bidirectional contrastive learning. GLoRIA~\cite{huang2021gloria} explored the fine-grained information contained in the image and report, proposing a framework for learning both global and regional representations of two modalities. 
MedCLIP~\cite{wang2022medclip} proposed one decoupled multimodal contrastive learning framework based on CLIP to scale the usable training data from two distinct sources.
BioVIL~\cite{boecking2022making} focused on the representation of radiological reports and proposed a radiology-specific text encoder along with the classical VLP paradigm. CheXzero~\cite{tiu2022expert} retrained a pre-trained CLIP model~\cite{radford2021learning} on the CXR data and showed considerable improvement. MedKLIP~\cite{wu2023medklip} extracted entities from reports and converted them to the medical-specific knowledge descriptions, which enhanced the model reasoning ability. KAD~\cite{zhang2023knowledge} built up a medical knowledge graph to fine-tune text encoder and performed the image-text contrastive learning with paired chest X-ray pairs, showing state-of-the-art capability on common benchmarks. However, all these works mentioned above are either pre-trained on a single dataset from the identical source or on dual sources like MedCLIP~\cite{wang2022medclip}, which overlook the non-negligible heterogeneity problem in their corresponding application on the multi-source CXR data.

\section{Methodology}\label{sec:method}

In this section, we first present the problem formulation and the motivation of our study. Then, we will introduce the basic model design that consists of modality-specific backbones and the MoE-QN module. Finally, we provide detailed descriptions and analysis of the Conquer-and-Divide pre-training stages.

\subsection{Problem Formulation}
Assuming that we have a training set of \textit{N} samples collected from the multiple sources, $\mathcal{D} = \{(x_i, Y_i, l_i, t_i)\}_{i=1}^N$, where $x_i$ denotes the CXR image, $Y_i$ is the label set indicating what diseases are found for $x_i$, $l_i$ means the source identity, and $t_i$ denotes the report of $x_i$. Note that, for those samples that have no reports, we convert $Y_i$ to $t_i$ correspondingly. Our goal is to train a vision-language pre-training model on the given multi-source data $\mathcal{D}$, which can accurately diagnose the chest diseases. Specifically, in the inference phase, the model can estimate the diseases in the given set for any CXR image. Here, we would like to clarify that our method adheres to the vision-language pre-training paradigm instead of transition supervised learning paradigm, since it inherently extracts representation vectors from images and text data for contrastive pre-training and can perform to open-set and zero-shot evaluations. Conventional supervised learning actually cannot perform such evaluations due to the rigid dimension issue about the prediction as well as the generalization dilemma towards the semantics of new classes.

\subsection{Motivation}
Generally, scaling up CXR samples for pre-training should be useful to improve generalization as discussed in the previous sections. However, we should point out that the resulting multi-source samples, on one hand, enjoy the better diversity \textit{w.r.t.} training samples and label space, yet on the other hand, suffer from the non-negligible source heterogeneity issue as shown in Fig.~\ref{motivation}. Physically, the locales and time frames of source samples can be quite diverse, as ChestX-ray14~\cite{wang2017chestx} contains the samples captured from 1992 to 2015 in the U.S. while VinDr-CXR~\cite{nguyen2022vindr} consists of CXRs in Vietnam from 2018 to 2020.  Even though CXRs might appear indistinguishable to the human eye, the diagnostic models can actually respond very differently in the face of some imperceptive factors like the dosage of X-ray used and the quality of the imaging instrument~\cite{rammuni2022effective}.
From the t-SNE visualization in Fig.~\ref{motivation}, we can find that the distributions of various sources exhibit the distinct disparities. 
\begin{figure}[t!]
\centerline{\includegraphics[width=\linewidth]{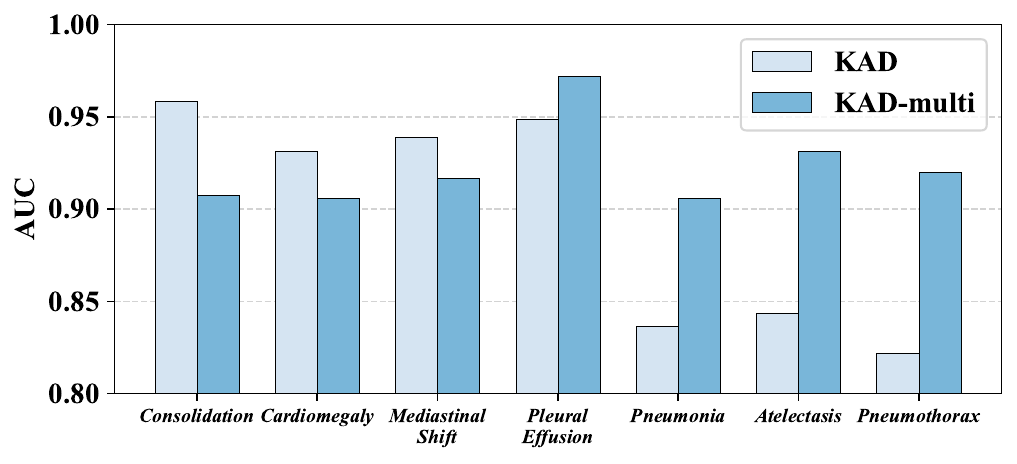}}
\caption{Inconsistent improvement is achieved when comparing KAD and KAD-Multi in terms of the AUC metric, when scaling up the training data by multiple sources and evaluating on the VinDr-CXR test set.}
\label{motivation_2}
\end{figure}
To further under the influence of the source heterogeneity, we implement the straightforward pre-training on the hybrid multi-source data and compare its performance with the pre-training on single source. Specifically, we adopt the current state-of-the-art (SOTA) \emph{KAD}~\cite{zhang2023knowledge} to conduct the pre-retraining on MIMIC-CXR dataset, and compare with the pre-training on the multiple sources (termed as \emph{KAD-multi}). As shown in Fig.~\ref{motivation_2}, KAD-multi does not achieve the consistent improvement, and on some certain diseases, KAD-multi even significantly lags behind the vanilla KAD.

With this observation, we re-think the early VLP paradigm that is naively applied in scaling up CXR data and overlooks the source heterogeneity issue. To handle this dilemma, we actually should allow the pre-training to capture the multi-source common patterns and simultaneously can maintain the source-specific patterns, which motivates us to incorporate the philosophy of ``Conquer" and ``Divide" design for pre-training.

\subsection{Conquer-and-Divide Pre-training}\label{subsec:twostage}
In this part, we describe the proposed Conquer-and-Divide pre-training framework, which includes the model architecture, the loss design and the training schedules, detailed as follows.

\subsubsection{The Model Architecture}
We follow the prevalent vision-language pre-training paradigm with the proper tailored design to train a diagnosis model, which consists of two modality-specific encoders and one modality-interaction module. Note that, this is different from the classical supervised framework that directly maps the input image space to the label space. The merits are tri-fold: First, we can inject more knowledge into the label space via the textual encoder, which is richer in semantics than the naive one/multi-hot label vector. The second is that we can incorporate the prior knowledge to promote the learning efficiency of vision encoder, when the textual description for medical images is available, e.g., the report information of samples from MIMIC. Finally, VLP allows us to achieve better generalization for open-set categories by means of the powered textual encoder. In the following, we concretely describe the architecture design of each component in our UniChest.  

Given a sample $(x_i, Y_i, l_i, t_i)$, we take the ResNet-50 as the visual backbone to encode $x_i$ and adopt the output of the $4$-th residual block as the image representation~\cite{wu2023medklip}, denoting as $\Phi_{\text{image}}(\cdot)$. For the report information $t_i$, 
RadGraph~\cite{jain2021radgraph} is used to extract key entities and filter the irrelevant words. When reports containing more than one sentence, we extract entities sentence by sentence, and use [SEP] token as the separation between different entities\footnote{For CXR samples without reports, we use their corresponding labels as the input entity and also take [SEP] token to separate the multiple labels.}. After the entity extraction, we use PubMedBERT~\cite{gu2021domain} pre-trained on Unified Medical Language System~(UMLS) data~\cite{bodenreider2004unified} as textual encoder backbone $\Phi_{\text{text}}$ for generating text representation~\cite{zhang2023knowledge}.
For clarity, we summarize the modality-specific encoding process as below,
\begin{equation} \label{eq:encoding}
\boldsymbol{I}_i = \Phi_{\text{image}}(x_i) \in \mathbb{R}^{h \times w \times d}, \quad
\boldsymbol{T}_i = \Phi_{\text{text}}(t_i) \in \mathbb{R}^{d}.
\end{equation}

For the disease prediction, we introduce a MoE-QN Module, namely, mixture of query networks, where each query network consists of a few transformer decoder layers~(4-layers in default). The MoE-QN module plays the role of overcoming the source heterogeneity issues by squeezing the source-specific patterns into different query networks, but its training should carefully follow our Conquer-and-Divide schedules, which will be discussed in the subsequent sections. For each query network, we take the fine-grained visual representation $\boldsymbol{I}_i$ as Key and Value, and utilize the textual representation of the disease set $Y_i = \{Y_{i1}, ..., Y_{ic}\}$ (encoded by $\Phi_{\text{text}}(\cdot)$ similarly as in Eq.~\eqref{eq:encoding}) as Query. The output from the sequential transformer decoder will be fed to one MLP layer to obtain the prediction of each Query Network,
\begin{equation}
{s_{i}^{k} = \Phi_{\text{QN}}^{k}(\boldsymbol{I}_i, \Phi_{\text{text}}(Y_i)) \in \mathbb{R}^{c},~k=1,...,K,}
\label{eq:formu}
\end{equation}
where $c$ is the total class number of all sources and $K$ is the number of query networks. Then, we transform $\{s_{i}^k\}_{k=1}^{K}$ of all query networks to the total prediction by an automatic linear combination, which will be described at the ``Divide" stage.

\begin{figure*}[t!]
\centerline{\includegraphics[width=\linewidth]{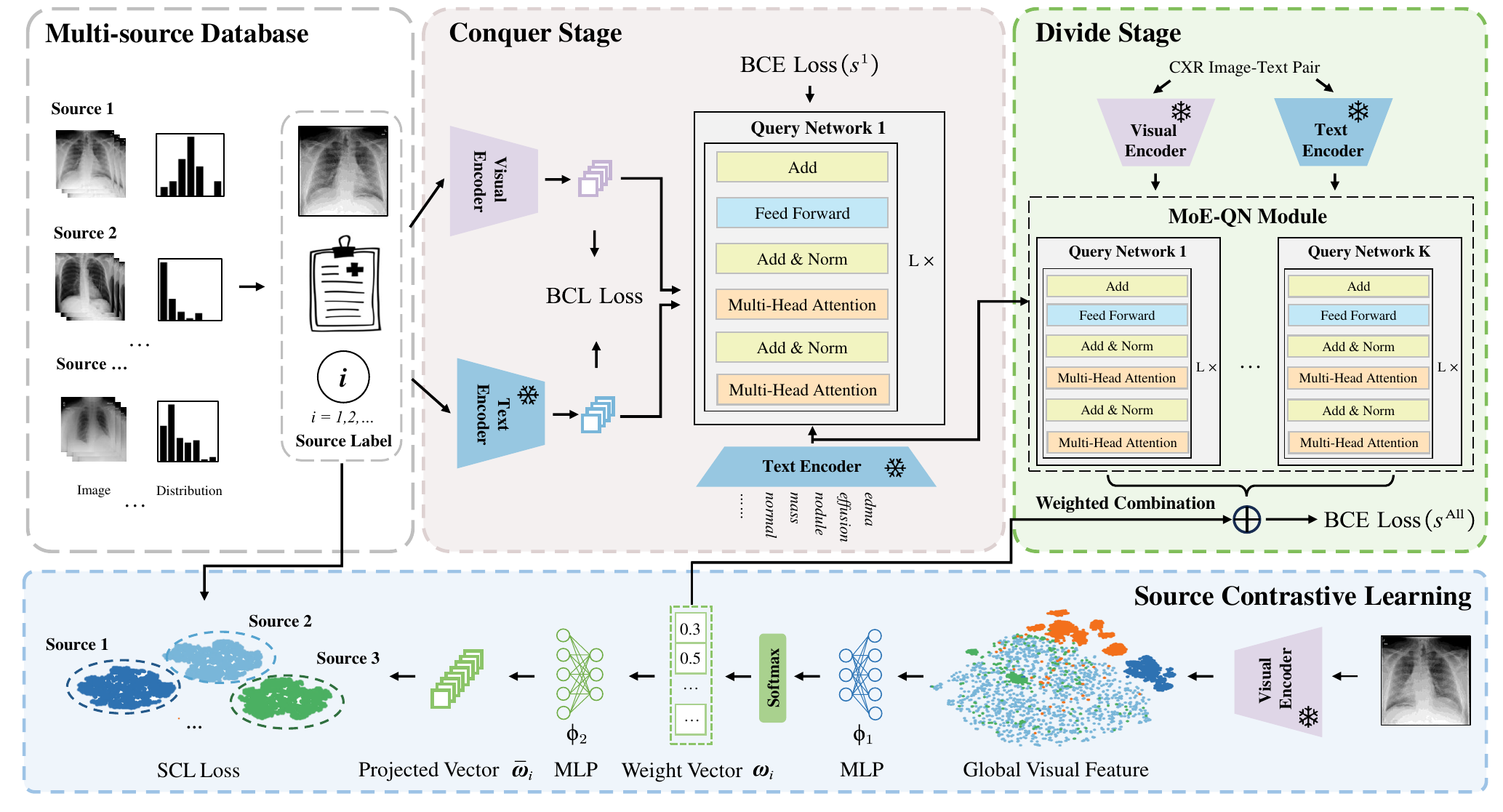}}
\caption{{The framework of UniChest, which consists of two training stages. During the ``Conquer" stage, two modality encoders first project visual and textual representations into the common space with alignment, then feed them into the first transformer query networks for prediction. The multi-source common patterns are learnt as much as possible at this stage. During the ``Divide" stage, we freeze the modality encoders and squeeze the source-specific patterns via the MoE-QN module with the guidance of the enhanced supervised loss and the source contrastive learning.}
}
\label{architecture}
\end{figure*}

\subsubsection{The Training at the ``Conquer" Stage} As discussed in the earlier sections, the source heterogeneity issue can be the core bottleneck to deteriorate the performance of pre-training for the consistent improvement for each source task. Therefore, at this ``Conquer" stage, we first encourage the model to capture multi-source common patterns as many as possible {following the prevalent vision-language pre-training framework like KAD~\cite{zhang2023knowledge}}. The training loss of this stage involves two parts, the image-text bidirectional contrastive loss $\mathcal{L}_{\text{BCL}}$ and the specially enhanced multi-label cross-entropy loss $\mathcal{L}_{\text{BCE}}$ (see Section IV.B for its implementation detail) \textit{w.r.t.} the \emph{first} query network prediction. Taking a batch of $M$ samples as an example, $\mathcal{L}_{\text{BCL}}$ emphasizes the alignment between the global visual representation and the textual representation, which could be formulated as follows, 
\begin{equation}
\mathcal{L}_{\text{BCL}}= -\log \frac{{\rm e}^{\bar{\boldsymbol{I}}_i^\top \boldsymbol{T}_i/\tau}}{\sum_{m=1}^{M} {\rm e}^{\bar{\boldsymbol{I}}_i^\top {\boldsymbol{T}}_m/\tau}} - \log \frac{{\rm e}^{\boldsymbol{T}_i^\top \bar{\boldsymbol{I}}_i/\tau}}{\sum_{m=1}^{M} {\rm e}^{\boldsymbol{T}_i^\top \bar{\boldsymbol{I}}_m/\tau}}
\end{equation}
where $\bar{\boldsymbol{I}}_i$ is the average pooling of $\boldsymbol{I}_i$ along the first two dimensions so that we reduce it to the same dimension of $\boldsymbol{T}_i$, and $\tau$ is temperature with the default setting 1.0 following~\cite{chen2020simple}. Regarding $\mathcal{L}_{\text{BCE}}$, we only enforce the supervision on the \emph{first} query network and compute the enhanced multi-label cross-entropy loss at this stage. Note that, samples in some domains may have smaller label space than the union label space of all sources. In this case, we neglect the computation between prediction and missing classes of these samples, even if it may belong to them (but are unobserved). Finally, we sum up $\mathcal{L}_{\text{BCL}}$ and $\mathcal{L}_{\text{BCE}}$ (only involves the first query network) as the overall loss for each mini-batch at the ``Conquer" stage as follows
\begin{equation}\label{eq:conquer}
\mathcal{L}_{\text{{Conquer}}}(s^1) = \mathcal{L}_{\text{BCL}} + \mathcal{L}_{\text{BCE}}(p, Y)\bigg|_{p=s^1}.
\end{equation}

\begin{table*}[t]
\footnotesize
\centering
\caption{
Comparison between UniChest and single-source pre-training with fine-tuning baselines on ChestX-ray14. AUC scores of 14 classes and the aAUC score are listed. Each label is abbreviated, which is aligned with labels in the left panel of Fig.~\ref{radar}.}

\begin{tabular}{c|c|cccccccccccccc}
\toprule[1.5pt]

  Method & Mean & Ate. & Car. & Eff. & Inf. & Mas. & Nod. & Pne. & Pne. & Con. & Ede. & Emp. & Fib. & Thi. & Her. \\

\midrule

ConVIRT & 0.808 & 0.771 & 0.867 & 0.825 & 0.703 & 0.818 & 0.761 & 0.722 & 0.857 & 0.747 & 0.854 & 0.901 & 0.809 & 0.771 & 0.909 \\
GLoRIA & 0.800 & 0.760 & 0.855 & 0.818 & 0.700 & 0.814 & 0.749 & 0.715 & 0.828 & 0.739 & 0.832 & 0.887 & 0.813 & 0.767 & 0.921 \\
BioViL & 0.800 & 0.765 & 0.871 & 0.824 & 0.697 & 0.819 & 0.752 & 0.710 & 0.845 & 0.742 & 0.842 & 0.871 & 0.821 & 0.759 & 0.888\\
MedKLIP & 0.801 & 0.764 & 0.849 & 0.823 & 0.697 & 0.820 & 0.747 & 0.712 & 0.839 & 0.751 & 0.848 & 0.879 & 0.817 & 0.777 & 0.892 \\
KAD  & 0.825 & 0.785 & 0.897 & 0.840 & 0.713 & 0.836 & 0.771 & 0.740 & 0.874 & 0.753 & 0.860 & 0.916 & 0.829 & 0.778 & 0.961 \\ 

\midrule

\textbf{UniChest}  & \textbf{0.858} & \textbf{0.823} & \textbf{0.922} & \textbf{0.861} & \textbf{0.739} & \textbf{0.850} & \textbf{0.778} & \textbf{0.777} & \textbf{0.933} & \textbf{0.795} & \textbf{0.893} & \textbf{0.958} & \textbf{0.877} & \textbf{0.837} & \textbf{0.977} \\ 

\bottomrule[1.5pt]
\end{tabular}
\label{method_discussion}
\end{table*}

\begin{table*}[t]
\footnotesize
\centering
\caption{
Comparison with other state-of-the-art methods on fine-tuning classification task with different training data portions on SIIM-ACR Pneumothorax. FT denotes fine-tuning pre-trained model, ZS denotes zero-shot classification.}

\begin{tabular}{c|cccc|cccc|cccc}
\toprule[1.5pt]
Data Portion & \multicolumn{4}{c|}{1\%} & \multicolumn{4}{c|}{10\%} & \multicolumn{4}{c}{100\%} \\ 
Method        & aAUC$\uparrow$      & aF1$\uparrow$       & aACC$\uparrow$ & mAP$\uparrow$  & aAUC$\uparrow$      & aF1$\uparrow$       & aACC$\uparrow$ & mAP$\uparrow$  & aAUC$\uparrow$      & aF1$\uparrow$       & aACC$\uparrow$ &  mAP$\uparrow$   \\ \midrule
ConVIRT~(FT) & 0.7134 & 0.4651 & 0.6375 & 0.4224 & 0.7826 & 0.5346 & 0.7190 & 0.4640 & 0.8989 & 0.6858 & 0.8351 & 0.7039 \\ 
GLoRIA~(FT) & 0.7439 & 0.4948 & 0.7030 & 0.4595 & 0.8538 & 0.6091 & 0.7874 & 0.6250 & 0.9018 & 0.6898 & 0.8450 & 0.7269 \\ 
BioVIL~(FT) & 0.6947 & 0.4685 & 0.6244 & 0.3816 & 0.7775 & 0.5393 & 0.7372 & 0.5183 & 0.8688 & 0.6364 & 0.8004 & 0.6624 \\ 
MedKLIP~(FT) & 0.8527 & 0.5952 & 0.7925 & 0.5858 & 0.9071 & 0.6845 & 0.8389 & 0.6823 & 0.9194 & 0.7246 & 0.8689 & 0.7510 \\ 
KAD~(FT) & 0.8541 & 0.5810 & 0.8097 & 0.4822 & 0.9023 & 0.6498 & 0.8490 & 0.6506 & 0.9203 & 0.6788 & 0.8655 & 0.7079 \\ 

\midrule

\textbf{UniChest~(ZS)} & 0.9260 & 0.6928 & \textbf{0.8807} & 0.6929 & 0.9260 & 0.6928 & 0.8807 & 0.6929 & 0.9260 & 0.6928 & 0.8807 & 0.6929 \\ 
\textbf{UniChest~(FT)} & \textbf{0.9375} & \textbf{0.7182} & 0.8801 & \textbf{0.7326} & \textbf{0.9418} & \textbf{0.7314} & \textbf{0.8881} & \textbf{0.7508} & \textbf{0.9437} & \textbf{0.7254} & \textbf{0.8885} & \textbf{0.7696} \\

\bottomrule[1.5pt]
\end{tabular}
\label{method_discussion_FT}
\end{table*}

\subsubsection{The Training at the ``Divide" Stage} Previous stage follows the classical pre-training spirit, where we treat multiple sources equally, and expect that the model can sufficiently learn the multi-source common patterns. At the ``Divide" phase, we mainly explore to mediate the negative impact induced by the source heterogeneity. Concretely, on the basis of the warming-up during the ``Conquer" stage, we train all the query networks~(MoE-QN Module) with \emph{freezing} all the other parts of the model. In the MoE-QN Module, the new query networks introduced at the this stage are initialized from randomness.
Here, the multi-label cross-entropy loss will be implemented on the ensemble prediction $s^{\text{All}}$ from $\{s^k\}_{k=1}^K$, unlike using $s^1$ in Eq.~\eqref{eq:conquer}. The final prediction is characterized by the following equation
\begin{equation} \label{eq:all_pred}
s_i^{\text{All}} = \lambda
 s_{i}^1 + (1-\lambda
)  \sum_{k=2}^{K} {s_{i}^k} {\omega_{i}^k},
\end{equation}
where $\lambda$ is the hyperparameter to balance the training of the first query network and the remaining query networks, and $\omega_i^k$ is the learnable weight to summarize the contribution of the source-specific modules for the diagnosis prediction. Note that, the intuition behind Eq.~\eqref{eq:all_pred} is to inherit the training gains at the ``Conquer" stage by $\lambda$, and simultaneously squeeze the source-specific patterns to the other query networks by $\omega_i^k$.

\noindent\textbf{Source Contrastive Learning.} In Eq.~\eqref{eq:all_pred}, we introduce the learnable weights to incorporate the source-specific patterns into the newly introduced query networks during the ``Divide" Stage.
However, without any guidance, it is challenging to achieve this goal by optimization. Here, we introduce a source contrastive learning to promote the desire. Specifically, we transform the global visual representation $\bar{\boldsymbol{I}}_i$ into a $K-1$ simplex via one-layer MLP $\Phi_{1}(\cdot)$ and Softmax layer as the weight vector, namely, $\boldsymbol{\omega}_i=\Phi_{1}(\bar{\boldsymbol{I}}_i)\in \mathbb{R}^{K-1}$. Then, $\boldsymbol{\omega}_i$ is projected into a higher dimension space by another one-layer MLP $\Phi_{2}(\cdot)$, denoted as $\bar{\boldsymbol{\omega}}_i=\Phi_2(\boldsymbol{\omega}_i)$, which is the high-dim vector to perform a source contrastive learning with the guidance of source id $l$ in the following 
\begin{equation}\label{eq:scl}
{\mathcal{L}_{\text{SCL}}= -\log \frac{\sum_{l_j = l_i}^{j \neq i} {\rm e}^{ \bar{\boldsymbol{\omega}}_i^\top \bar{\boldsymbol{\omega}}_j/\tau}}{\sum_{m \neq i}^{M} {\rm e}^{\bar{\boldsymbol{\omega}}_i^\top \bar{\boldsymbol{\omega}}_m/\tau}} }.
\end{equation}
In Eq.~\eqref{eq:scl}, the numerator ${\sum_{l_j = l_i}^{j \neq i} {\rm e}^{ \bar{\boldsymbol{\omega}}_i^\top \bar{\boldsymbol{\omega}}_j/\tau}}$ computes the sum of exponential similarities of the high-dim weight vectors $\bar{\boldsymbol{\omega}}$ between sample $i$ and all other samples $j$ from the same source $l$, encouraging to pull the samples with the same source identify closer together. The denominator ${\sum_{m \neq i}^{M} {\rm e}^{\bar{\boldsymbol{\omega}}_i^\top \bar{\boldsymbol{\omega}}_m/\tau}}$ calculates the sum of exponential similarities of the high-dim weight vectors between sample $i$ and all other samples in the batch, providing a normalization factor to ensure stability of the loss. Generally,
above contrastive learning helps us to learn the similar weight vector for the samples from the same source, which makes the source-specific patterns learned by the similar query networks. Under this mechanism, we naturally learn an automatic optimal assignment for the remaining query networks to overcome the source heterogeneity issue. Then, with Eq.~\eqref{eq:all_pred} and Eq.~\eqref{eq:scl}, the overall loss at the ``Divide" stage can be formulated as
\begin{equation}
\mathcal{L}_{\text{{Divide}}} = \mathcal{L}_{\text{SCL}} + \mathcal{L}_{\text{Conquer}}(s^{\text{All}})
\end{equation}

\begin{table*}
\centering
\caption{Statistics of datasets for pre-training and downstream.}
\label{dataset}
\setlength{\tabcolsep}{9.4pt}
\begin{tabular}{c|l|rrlll}
\toprule[1.5pt]
Purpose & Dataset & \# Samples &  \# Diseases & Region & Collection Institution & Time Scope\\
\midrule
\multirow{5}{*}{Pre-training}& MIMIC-CXR~\cite{johnson2019mimic} & 348900 & 41 & Northeast USA & MIT & 2011-2016\\
&ChestX-ray14~\cite{wang2017chestx} & 98637  & 14 & Northeast USA & NIH & 1992-2015\\
&CheXpert~\cite{irvin2019chexpert} & 223414  & 14 & Western USA & Stanford Hospital & 2002-2017\\
&VinDr-CXR~\cite{nguyen2022vindr} & 15000  & 28 & Vietnam & VinBrain & 2018-2020\\

& Multi-CXR & 685951  & 60 & USA \& Vietnam & MIT, \textit{etc.} & 1992-2020 \\
\midrule
\multirow{5}{*}{Downstream}& Shenzhen~\cite{jaeger2014two} & 662 & 1 & Southeast China & Guangdong Medical College & Sep. 2012\\
& Open-I~\cite{demner2016preparing, moon2022multi} & 3547 & 14 & USA & NLM & / \\
& SIIM-ACR~\cite{kaggle-siim, wu2023medklip} & 2135 & 1 & Northeast USA & Kaggle & / \\
& PadChest~\cite{demner2016preparing} & 39053 & 193 & Spain & University of Alicante & 2009-2017\\
& ChestX-Det10~\cite{liu2020chestx} & 542 & / & Northeast USA & Deepwise &  1992-2015\\
\bottomrule[1.5pt]
\end{tabular}
\end{table*}

\subsubsection{Difference from the pre-training with fine-tuning}
The proposed Conquer-and-Divide pre-training is intrinsically different from the ordinary fine-tuning after pre-training, although both of them are a two-stage process. First, the two-stage training of UniChest is towards all multi-source data, while the fine-tuning after pre-training is narrowed down to a single-source scenario. When the target dataset's corresponding training set is seen during multi-source pre-training, this leads to the distinction in the generalization ability of two frameworks, where UniChest is significantly stronger than pre-training with fine-tuning as supported by Table~\ref{method_discussion}.
Second, the ``Divide" stage is to properly mediate the heterogeneity patterns in a broad sense, which does not mean multi-source pre-training contradicts with the conventional pre-training and fine-tuning diagram. Conversely, the downstream task can still apply the ordinary fine-tuning ways to achieve fast adaptation for unseen sources. For instance, we fine-tune the pre-trained UniChest on SIIM-ACR Pneumothorax~\cite{kaggle-siim} with different training data ratios following MedKLIP~\cite{wu2023medklip}. Overall, UniChest's zero-shot capability is already comparable to the 100\% fine-tuned performance of other models and the diagnostic performance is further enhanced after further fine-tuning as shown in Table~\ref{method_discussion_FT}.
Finally, when it comes to the subsequent experimental comparison, we mainly focus on the zero-shot performance of pre-training, instead of the performance by the fine-tuning, which follows the prevalent pre-training spirit that tries to avoid the expensive tuning cost for downstream tasks as much as possible.

\section{Experiments} \label{experiment}
In this section, we will introduce the datasets for pre-training and downstream tasks, evaluation metrics and implementation details will also be detailedly described. At last, we will present the experimental results of our proposed method compared with other baselines.

\subsection{Datasets}
We combine some common CXR datasets containing \textbf{MIMIC-CXR}~\cite{johnson2019mimic}, \textbf{ChestX-ray14}~\cite{wang2017chestx}, \textbf{CheXpert}~\cite{irvin2019chexpert} and \textbf{VinDr-CXR}~\cite{nguyen2022vindr} for pre-training, namely \textbf{Multi-CXR}, which yields a grand total of 685,951 Chest X-Ray images.

For evaluating the generalization performance of the pre-trained model, we adopt \textbf{Shenzhen}~\cite{jaeger2014two}, \textbf{Open-I}~\cite{demner2016preparing}, \textbf{SIIM-ACR Pneumothorax}~\cite{kaggle-siim} and \textbf{PadChest}~\cite{demner2016preparing} for zero-shot classification and \textbf{ChestX-Det10}~\cite{liu2020chestx} which is the subset of ChestX-ray14 for intuitive lesion grounding. More detailed statistics of pre-training and downstream datasets can be found in Table~\ref{dataset}. We can observe that significant diversity exists in the collection time, population distribution, disease category coverage and sample scales among different datasets.

\subsection{Implementation Details}

The pre-training process of UniChest is conducted on a single NVIDIA A100 GPU for 30 epochs in the first ``Conquer'' stage and 20 epochs in the second ``Divide'' stage. The starting checkpoint for the second stage is the one that performed the best on the ChestX-ray14 validation set. {The number of transformer decoder layers of each query network in MoE-QN Module is set to 4.} The hyperparameter $\lambda$ which weights the first query network during the ``Divide'' stage is set to be $0.5$ and the total number of query network $K$ is $4$ in default. {The dimension of $\bar{\boldsymbol{\omega}}$ which is the projection of the MLP layer $\Phi_2(\cdot)$ is set to be 32 during source contrastive learning. The temperature parameter $\tau$ in Eq.~\eqref{eq:scl} is set to 1.0.}
We set the training batch size as 32 and resize input images as 512 $\times$ 512.
We adopt the AdamW optimizer in conjunction with the cosine annealing scheduler for managing the learning rate, where the initial learning rate is $1 \times 10^{-5}$. 
To perform data augmentation, we additionally utilize the Fourier amplitude mixup method, which has been demonstrated to be effective for medical imaging data~\cite{semidg2022isbi, feddg2021cvpr}. 
Besides, we utilize the ASL loss~\cite{ridnik2021asymmetric} as $\mathcal{L}_{\text{BCE}}$ {to promote balanced training}. These techniques have demonstrated promising outcomes in the analysis of CXR data.
{For MedKLIP-multi and KAD-multi, we strictly follow the training hyper-parameters and implementation details described in their original papers and official codes. Their pre-training dataset is also Multi-CXR that is identical to UniChest. In MedKLIP-multi, for samples without location annotations~(excluding MIMIC-CXR), we mask the calculation of location contrastive loss directly.}

\subsection{Evaluation Metrics}
Some common metrics for multi-label classification are adopted to evaluate the model performance, {\em i.e.}, area under curve~(AUC), F1 score, accuracy~(ACC) and average precision~(AP) for each category and their average value, namely, average AUC~(aAUC), average F1~(aF1), average accuracy~(aACC) and mean average precision~(mAP), for comprehensive comparison. Following the strategy of MedKLIP~\cite{wu2023medklip}, the final binary threshold of each category prediction is the value when the maximum F1 score is achieved and ACC metric also adopts this threshold.

\subsection{Baselines}
We consider a wide range of baselines for CXR pre-training, including \textbf{ConVIRT}~\cite{zhang2020contrastive}, \textbf{GLoRIA}~\cite{huang2021gloria}, \textbf{MedCLIP}~\cite{wang2022medclip}, \textbf{BioVIL}~\cite{boecking2022making}, \textbf{CheXzero}~\cite{tiu2022expert}, \textbf{MedKLIP}~\cite{wu2023medklip}, and \textbf{KAD}~\cite{zhang2023knowledge}.

\textbf{ConVIRT} trains two modality-specific encoders by bidirectional contrastive loss to learn visual representations.~\textbf{GLoRIA} utilizes both global and fine-grained features for medical VLP. \textbf{MedCLIP}~\cite{wang2022medclip} trains one CLIP-based framework on two CXR datasets including MIMIC-CXR and CheXpert. \textbf{BioVIL} proposes a radiology-specific text encoder for the subsequent classical pipeline of VLP. \textbf{CheXzero} retrains one CLIP model with a corpus of the medical domain. \textbf{MedKLIP} designs one novel entity extraction and transition module to inject domain-specific knowledge into the process of VLP. \textbf{KAD} incorporates a medical knowledge graph to further improve the capability of current VLP models in Chest X-Ray, showing SOTA performance on some common public~benchmarks.

\begin{table*}[t]
\footnotesize
\centering
\caption{Comparison of UniChest with baselines on in-domain classification. Four metrics including aAUC, aF1, aACC and mAP scores are reported. For all datasets, the metrics all refer to the macro average on all diseases.}
\setlength{\tabcolsep}{6.8pt}
\begin{tabular}{c|cccc|cccc|cccc}
\toprule[1.5pt]
Dataset & \multicolumn{4}{c|}{ChestX-ray14} & \multicolumn{4}{c|}{CheXpert} & \multicolumn{4}{c}{VinDr-CXR} \\ 
Method        & aAUC$\uparrow$      & aF1$\uparrow$       & aACC$\uparrow$ & mAP$\uparrow$  & aAUC$\uparrow$      & aF1$\uparrow$       & aACC$\uparrow$ & mAP$\uparrow$  & aAUC$\uparrow$      & aF1$\uparrow$       & aACC$\uparrow$ &  mAP$\uparrow$   \\ \midrule
ConVIRT & 0.5804 & 0.1623 & 0.5385 & 0.1008 & 0.6640 & 0.3609 & 0.7232 & 0.2990 & 0.6803 & 0.1748 & 0.8233 & 0.1129 \\
GLoRIA & 0.6099 & 0.1803 & 0.5616 & 0.1220 & 0.6889 & 0.4094 & 0.7472 & 0.3580 & 0.6775 & 0.1953 & 0.7796 & 0.1380\\
MedCLIP & 0.6696 & 0.2061 & 0.7001 & 0.1424 & 0.7228 & 0.4149 & 0.7719 & 0.3613 & 0.6958 & 0.2079 & 0.8163 & 0.1503 \\
BioViL & 0.5766 & 0.1561 & 0.5842 & 0.0948 & 0.5769 & 0.3080 & 0.6446 & 0.2157 & 0.5856 & 0.1507 & 0.7726 & 0.1048\\ 
CheXzero & 0.6872 & 0.2205 & 0.7698 & 0.1597 & 0.7537 & 0.4488 & 0.8136 & 0.4069 & 0.7407 & 0.2354 & 0.8562 & 0.1713\\
MedKLIP & 0.7233 & 0.2541 & 0.7946 & 0.1860 & 0.8389 & 0.5385 & 0.8672 & 0.5130 & 0.7770 & 0.2433 & 0.8517 & 0.1897\\
KAD & 0.7933 & 0.3363 & 0.8639 & 0.2746 & 0.8165 & 0.5467 & 0.7835 & 0.5200 & 0.7599 & 0.2960 & 0.8724 & 0.2315\\
\midrule
KAD-CXR14 & 0.8380 & 0.4006 & 0.8913 & 0.3401 & 0.7455 & 0.4296 & 0.7658 & 0.3901 & 0.7711 & 0.2399 & 0.8410 & 0.1740 \\
KAD-CXP &  0.7014 & 0.2444 & 0.7744 & 0.1767 & 0.8804 & 0.5964 & 0.8846 & 0.5606 & 0.7382 & 0.2256 & 0.8101 & 0.1619 \\
KAD-VC & 0.6372 & 0.1881 & 0.6776 & 0.1275 & 0.7228 & 0.4161 & 0.7811 & 0.3285 & 0.8621 & 0.3839 & 0.9418 & 0.3255 \\
\midrule
MedKLIP-multi & 0.7915 & 0.2994 & 0.8691 & 0.2313 & 0.7010 & 0.4244 & 0.7395 & 0.3410 & 0.7879 & 0.2802 & 0.9167 & 0.2196\\
KAD-multi & 0.8431 & 0.4077 & 0.8905 & 0.3457 & 0.8819 & 0.5912 & 0.8713 & 0.5626 & 0.8716 & 0.3698 & 0.9390 & 0.3202 \\
\midrule

\textbf{UniChest} & \textbf{0.8584} & \textbf{0.4293} & \textbf{0.8999} & \textbf{0.3797} & \textbf{0.9005} & \textbf{0.6446} & \textbf{0.8912} & \textbf{0.6328} & \textbf{0.8807} & \textbf{0.4028} & \textbf{0.9501} & \textbf{0.3520}\\ 
\bottomrule[1.5pt]
\end{tabular}
\label{main_2}
\end{table*}

\begin{table*}[t]
\footnotesize
\centering
\caption{Comparison of UniChest with baselines on zero-shot classification. Four metrics including aAUC, aF1, aACC and mAP scores are reported. For single-labeled dataset, we report AUC, F1, ACC and AP scores.}
\setlength{\tabcolsep}{7pt}
\begin{tabular}{c|cccc|cccc|cccc}
\toprule[1.5pt]
Dataset & \multicolumn{4}{c|}{Shenzhen} & \multicolumn{4}{c|}{Open-I} & \multicolumn{4}{c}{SIIM-ACR Pneumothorax} \\ 
Method        & AUC$\uparrow$      & F1$\uparrow$       & ACC$\uparrow$ & AP$\uparrow$  & aAUC$\uparrow$      & aF1$\uparrow$       & aACC$\uparrow$ &  mAP$\uparrow$   & AUC$\uparrow$      & F1$\uparrow$       & ACC$\uparrow$ & AP$\uparrow$ \\ 
\midrule
ConVIRT & 0.7166 & 0.6908 & 0.6375 & 0.7619 & 0.6265 & 0.1855 & 0.7504 & 0.1115 & 0.6356 & 0.4322 & 0.4778 & 0.2948 \\
GLoRIA & 0.5831 & 0.6733 & 0.5060 & 0.6001 & 0.6449 & 0.2218 & 0.8134 & 0.1535 & 0.5342 & 0.3770 & 0.4047 & 0.2084\\
MedCLIP & 0.6545 & 0.6933 & 0.5160 & 0.6321 & 0.6609 & 0.2422 & 0.8406 & 0.1593 & 0.6822 & 0.4611 & 0.5564 & 0.3159 \\
BioViL & 0.6227 & 0.6872 & 0.5544 & 0.6174 & 0.5669 & 0.1515 & 0.6966 & 0.0941  & 0.4678 & 0.3109 & 0.1836 & 0.1761\\ 
CheXzero & 0.7801 & 0.7302 & 0.6692 & 0.7969 & 0.6887 & 0.2429 & 0.8689 & 0.1850 & 0.6879 & 0.4722 & 0.5588 & 0.3240\\
MedKLIP & 0.7909 & 0.7741 & 0.6906 & 0.7986 & 0.7219 & 0.2897 & 0.9048 & 0.2119 & 0.8924 & 0.6833 & 0.8428 & 0.6869\\
KAD & 0.8856 & 0.8474 & 0.7893 & 0.8141 & 0.7346 & 0.3248 & 0.8808 & 0.2603 & 0.8967 & 0.6480 & 0.8385 & 0.6093 \\
\midrule
KAD-CXR14 & 0.8764 & 0.8037 & 0.8000 & 0.8947 & 0.6986 & 0.2676 & 0.8617 & 0.1856 & 0.9162 & 0.6561 & 0.8348 & 0.6872 \\
KAD-CXP & 0.7204 & 0.6793 & 0.6031 & 0.7611 & 0.7701 & 0.3314 & 0.9011 & 0.2631 & 0.9132 & 0.6554 & 0.8456 & 0.6771 \\
KAD-VC & 0.8832 & 0.8448 & 0.8578 & 0.9186 & 0.7012 & 0.2112 & 0.8468 & 0.1503 & 0.8002 & 0.4995 & 0.7661 & 0.4012 \\
\midrule

MedKLIP-multi & 0.8431 & 0.7780 & 0.7855 & 0.8817 & 0.6440 & 0.2270 & 0.7943 & 0.1469 & 0.8958 & 0.6888 & 0.8412 & 0.6853  \\
KAD-multi & 0.8794 & 0.8419 & 0.8547 & 0.9180 & 0.7736 & 0.3409 & 0.9065 & 0.2749 & 0.9016 & 0.6515 & 0.8395 & 0.5847 \\
\midrule

\textbf{UniChest} & \textbf{0.9499} & \textbf{0.8906} & \textbf{0.8913} & \textbf{0.9598} & \textbf{0.7830} & \textbf{0.3520} & \textbf{0.9104} & \textbf{0.2877} & \textbf{0.9260} & \textbf{0.6928} & \textbf{0.8807} & \textbf{0.6929}\\ 
\bottomrule[1.5pt]
\end{tabular}
\label{main_1}
\end{table*}

\subsection{Results}
In this part, we provide the results of in-domain evaluation on pre-training datasets and zero-shot evaluation on downstream datasets. Additionally, we showcase the lesion grounding capability through intuitive examples.

\subsubsection{In-domain evaluation on pre-training datasets}
\begin{figure*}[t!]
\centerline{\includegraphics[width=0.95\textwidth]{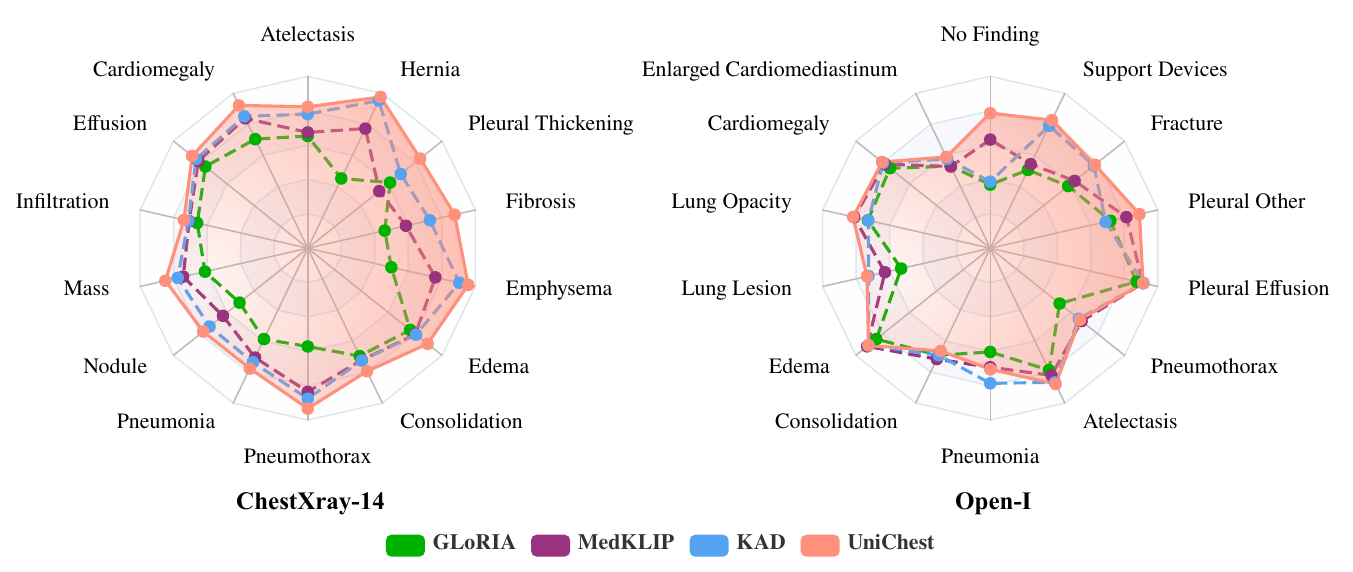}}
\caption{Per-category performance of different methods on ChestX-ray14~(left) and Open-I~(right). AUC scores of each category are displayed. 1 and 0 are adopted as the maximum and minimal values for each category in the radar chart.}
\label{radar}
\end{figure*}

We compare UniChest with other baselines on the test set of pre-training datasets in Table~\ref{main_2}. UniChest demonstrates a notable improvement in various evaluation metrics compared to the best baseline models. Specifically, for ChestX-ray14, UniChest shows a significant improvement of 6.51\%, 9.30\%, 3.60\% and 10.51\% respectively as to aAUC, aACC, aF1 and mAP than the best baseline. In terms of fine-grained classification for each category, UniChest demonstrates persistent gains as shown in Fig{.}~\ref{radar} (left). For CheXpert, UniChest outperforms 6.16\%, 9.79\%, 2.40\% and 11.28\%. In the case of VinDr-CXR, UniChest surpasses the best baselines by 12.08\%, 10.68\%, 7.77\% and 12.05\%. By introducing multi-source CXR datasets and the Conquer-and-Divide pre-training framework, the diagnosis ability of VLP model has been significantly enhanced.

To mitigate the impact of inconsistent training data, we also conduct variants of the baselines that utilize the same training data with UniChest. These variants include models trained on single-source data (\emph{i.e.,} KAD-CXR14, KAD-CXP, and KAD-VC in Table~\ref{main_2}) as well as models trained on multi-source data (\emph{i.e.,} MedKLIP-multi and KAD-multi). KAD-CXR14 outperforms KAD on ChestX-ray14 with improvements of 4.47\% in aAUC, 6.43\% in aACC, 2.74\% in aF1, and 6.55\% in mAP. However, it falls short compared to UniChest, which achieves better results by 2.04\%, 2.87\%, 0.86\%, and 3.96\% respectively. KAD-CXP and KAD-VC also demonstrate similar performance. Moreover, it is worth noting that while the single dataset-based KAD models may perform reasonably well on their corresponding datasets, their generalization capability to other datasets is limited. This is evident in the case of KAD-VC's performance on ChestX-ray14, where it may not generalize effectively. For the baseline variants that utilize multi-source data, MedKLIP-multi and KAD-multi, their overall performance exceeds that of their vanilla methods and achieves comparable or even better results than using only the corresponding single-source data. However, these two methods of direct data replenishment lag behind our proposed UniChest with a substantial margin. Generally speaking, the comparison with these variants validates the necessity and importance of developing a multi-source CXR foundational model and demonstrates the effectiveness of our framework design.

\begin{figure*}[t!]
\centering
\includegraphics[width=0.9\linewidth]{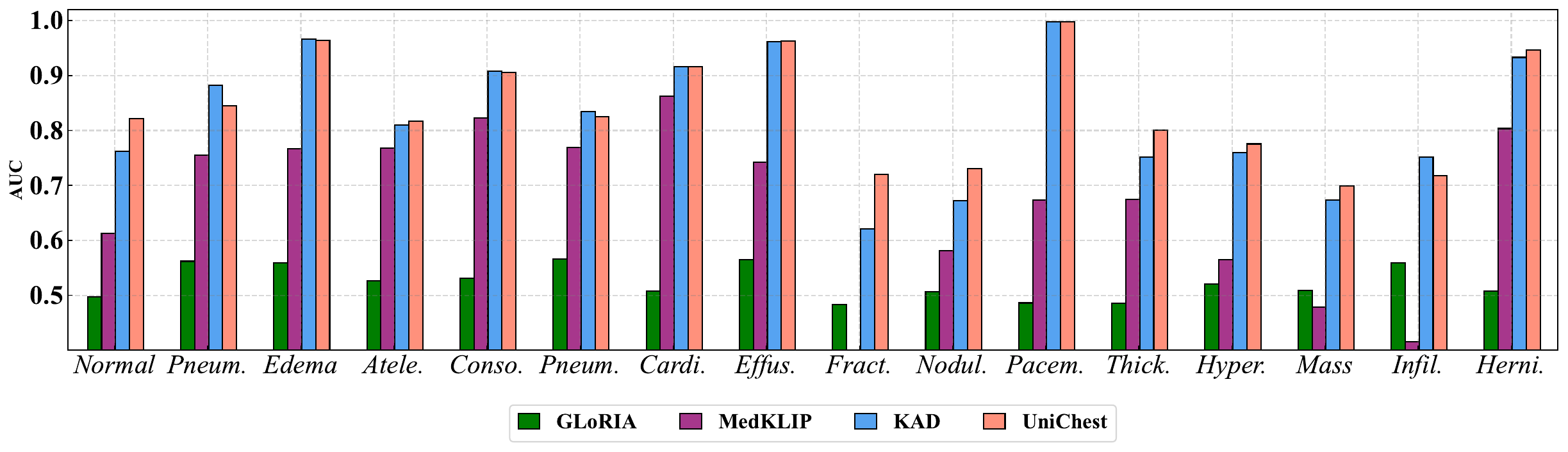}
\caption{Per-category performance of 16 \textit{seen} categories during pre-training in PadChest.}
\label{padchest_1}
\end{figure*}

\subsubsection{Zero-shot classification evaluation}
\begin{figure*}[t!]
\centering
\includegraphics[width=\linewidth]{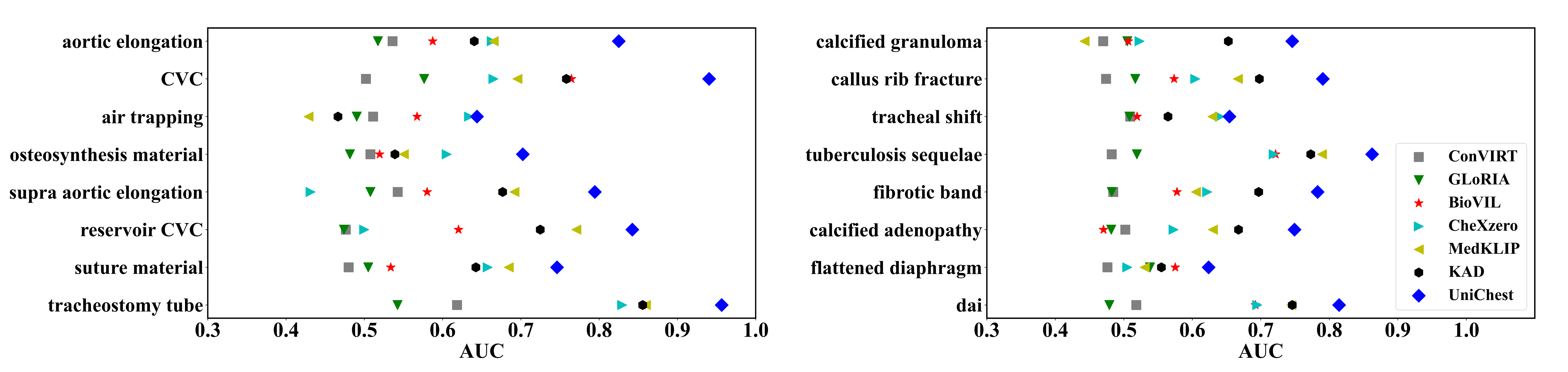}
\caption{Per-category performance of 16 randomly selected \textit{unseen} categories with more than 50 positive samples in PadChest.}
\label{padchest_2}
\end{figure*}

We conduct the zero-shot evaluation to assess the generalization ability of CXR pre-training models. The results in Table~\ref{main_1} demonstrate the superiority of UniChest across various datasets.
In the case of the Shenzhen dataset, UniChest surpasses the best baseline KAD by 6.43\%, 4.32\%, 10.20\% and 14.57\% respectively as to AUC, ACC, F1 and AP. 
For the SIIM-ACR Pneumothorax dataset, UniChest outperforms 2.93\%, 0.95\%, 3.79\% and 0.60\% respectively compared with the best baseline. 
As to Open-I dataset, UniChest achieves significant improvements by 4.84\%, 2.72\%, 0.56\% and 2.74\%. As indicated by the per-category evaluation in Fig.~\ref{radar} (right), UniChest achieves performance improvements across the majority of the categories. 
For instance, the classification ability is sharply consolidated of \textit{No Finding}, \textit{Pleural Other} and \textit{Support Devices} with a margin of over 3\%.
For PadChest, UniChest obtains an average AUC of 0.8403 for 16 \textit{seen} categories during pre-training, surpassing KAD by 1.51\%. Specifically, the diagnosis capability of 11 classes achieves SOTA as shown in Fig.~\ref{padchest_1}, among which the improvements of \textit{fracture}, \textit{nodule} and \textit{pleural thickening} are around or over 5\%.
For unseen pathologies in Fig.~\ref{padchest_2}, UniChest also showcases superior performance in the given 16 diseases, demonstrating its value in rare disease diagnosis.
Besides, it is worth noting that the multi-source data baseline variants (\emph{e.g.}, KAD-multi) exhibit relatively modest and inconsistent improvements compared to their vanilla methods and single-source data baseline variants (\emph{e.g.}, KAD-CXR14), which emphasizes the necessity of our method design in enhancing the model's generalization ability.
In summary, the results in Table~\ref{main_1} validate the significance of the Conquer-and-Divide pre-training framework for multi-source CXR samples, which indicates its potential in assisting clinical human diagnosis and highlights its unignorable value.

\subsubsection{Qualitative grounding visualization}
\begin{figure*}[t!]
\centerline{\includegraphics[width=\linewidth]{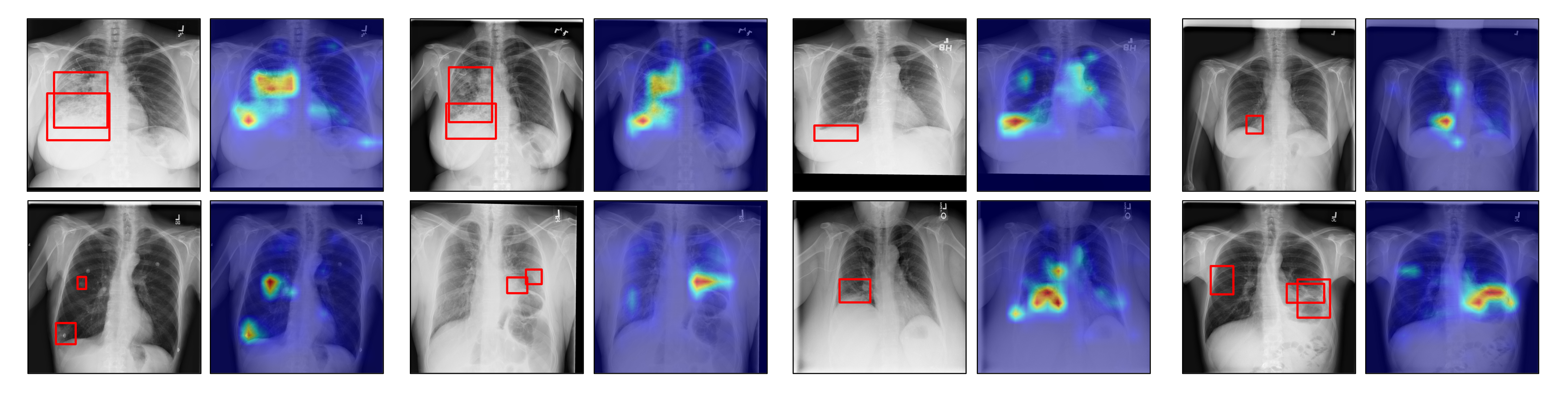}}
\caption{Some visualization examples of lesion grounding performance of UniChest on ChestX-Det10.
In the left CXR images, bounding boxes are abnormal areas manually annotated by the consensus of some board-certified radiologists. In the right heatmaps generated by UniChest, the color temperature in the heatmaps indicates the attention focus of the model, with higher temperatures indicating greater attention and suggesting a higher likelihood of abnormality in the corresponding region.}
\label{fig:grounding}
\end{figure*}

In Fig.~\ref{fig:grounding}, we present several examples of lesion grounding on ChestX-Det10 by UniChest. To provide an intuitive visualization, we generate spectrum heatmaps on the original CXR images based on the MoE-QN Module's regional cross-attention maps in transformer decoder layers. By comparing the model-detected lesions with the bounding boxes annotated by expert clinicians, we observe a strong alignment between model findings and the diagnoses made by experts, demonstrating the reasoning ability and interpretability of UniChest.

\begin{table*}[t]
\footnotesize
\centering
\caption{Ablation study on different stages of UniChest. Four metrics including aAUC, aF1, aACC and mAP scores are reported. For single-labeled dataset, we report AUC, F1, ACC and AP scores.}

\setlength{\tabcolsep}{5.2pt}

\begin{tabular}{c|cccc|cccc|cccc}
\toprule[1.5pt]
Dataset & \multicolumn{4}{c|}{Shenzhen} & \multicolumn{4}{c|}{SIIM-ACR Pneumothorax} & \multicolumn{4}{c}{ChestX-ray14} \\ 
Method        & AUC$\uparrow$      & F1$\uparrow$       & ACC$\uparrow$ & AP$\uparrow$  &  AUC$\uparrow$      & F1$\uparrow$       & ACC$\uparrow$ & AP$\uparrow$  & aAUC$\uparrow$      & aF1$\uparrow$       & aACC$\uparrow$ &  mAP$\uparrow$   \\ \midrule

KAD~\textit{(single-source SOTA)} & 0.8856 & 0.8474 & 0.7893 & 0.8141 & 0.8967 & 0.6833 & 0.8428 & 0.6869 & 0.7933 & 0.3363 & 0.8639 & 0.2746 \\

UniChest~(\textit{only Conquer stage}) & 0.9013 & 0.8558 & 0.8594 & 0.9313 & 0.9019 & 0.6450 & 0.8494 & 0.6601 & 0.8468 & 0.4143 & 0.8957 & 0.3543\\

UniChest~(\textit{with equal weights}) & 0.9463 & 0.8851 & 0.8838 & 0.9508 & 0.8687 & 0.5867 & 0.8234 & 0.5507 & 0.8528 & 0.4250 & 0.8971 & 0.3722 \\

\midrule

\textbf{UniChest} & \textbf{0.9499} & \textbf{0.8906} & \textbf{0.8913} & \textbf{0.9598} & \textbf{0.9260} & \textbf{0.6928} & \textbf{0.8807} & \textbf{0.6929} & \textbf{0.8584} & \textbf{0.4293} & \textbf{0.8999} & \textbf{0.3797} \\ 
\bottomrule[1.5pt]
\end{tabular}
\label{ablation_2}
\end{table*}

\section{Ablation Study} \label{sec:ablation}

\subsection{The Effectiveness of Conquer-and-Divide Framework}
\subsubsection{On ``Conquer" stage}
Data replenishment from multiple sources is an intractable issue mentioned in Section~\ref{sec:method}. We compare the impact of incorporating multi-source training data in the ``Conquer'' stage of our UniChest on the diagnostic performance, as displayed in the second row of Table~\ref{ablation_2}. It turns out that simply augmenting the training data with multiple sources does not consistently yield significant performance gains compared to using a single source, as shown in the first row of the table. According to Table~\ref{ablation_2}, explicit data replenishment improves the performance on ChestX-ray14 by 6.08\%, aligning with the training data distribution. However, the performance of the zero-shot evaluation varies between the two inference sets. For the Shenzhen dataset, the influence of the ``Conquer'' stage has a significantly positive effect, as all four metrics experience considerable improvement. In the case of SIIM-ACR Pneumothorax, the AUC and ACC values are comparable to previous state-of-the-art approaches, but the other two metrics decrease by approximately 3.26\%.

\subsubsection{On ``Divide" stage}
A comparison of the second and fourth rows of Table~\ref{ablation_2} reveals that UniChest significantly outperforms the ``Conquer'' stage in terms of zero-shot generalization by 4.86\% (AUC), 3.48\% (F1), 3.19\% (ACC), and 2.85\% (AP) for Shenzhen, and 2.41\% (AUC), 4.78\% (F1), 3.13\% (ACC), and 3.28\% (AP) for SIIM-ACR Pneumothorax. To further present the effectiveness of the soft-gating mechanism and $\mathcal{L}_{\text{SCL}}$ in the MoE-QN structure during the ``Divide'' stage, we present the results of removing the weight generation process and replacing it with equal weights in the third row of Table~\ref{ablation_2}. As shown on SIIM-ACR Pneumothorax, our mechanism significantly enhances zero-shot generalization. 
In addition to the performance improvement, it's also important to note that the introduction of the MoE-QN structure at the ``Divide'' stage does not significantly increase the computational cost compared to previous single-stage pre-training frameworks. Concretely, the increase in computational complexity is marginal, growing only from 44.3 GFLOPs in KAD to 45.5 GFLOPs in UniChest, which constitutes an increase of approximately 2.7\%.

\begin{figure*}[t!]
    \centering
    \begin{minipage}{0.24\textwidth}
        \subfigure[aAUC Score]{\includegraphics[width=\textwidth]{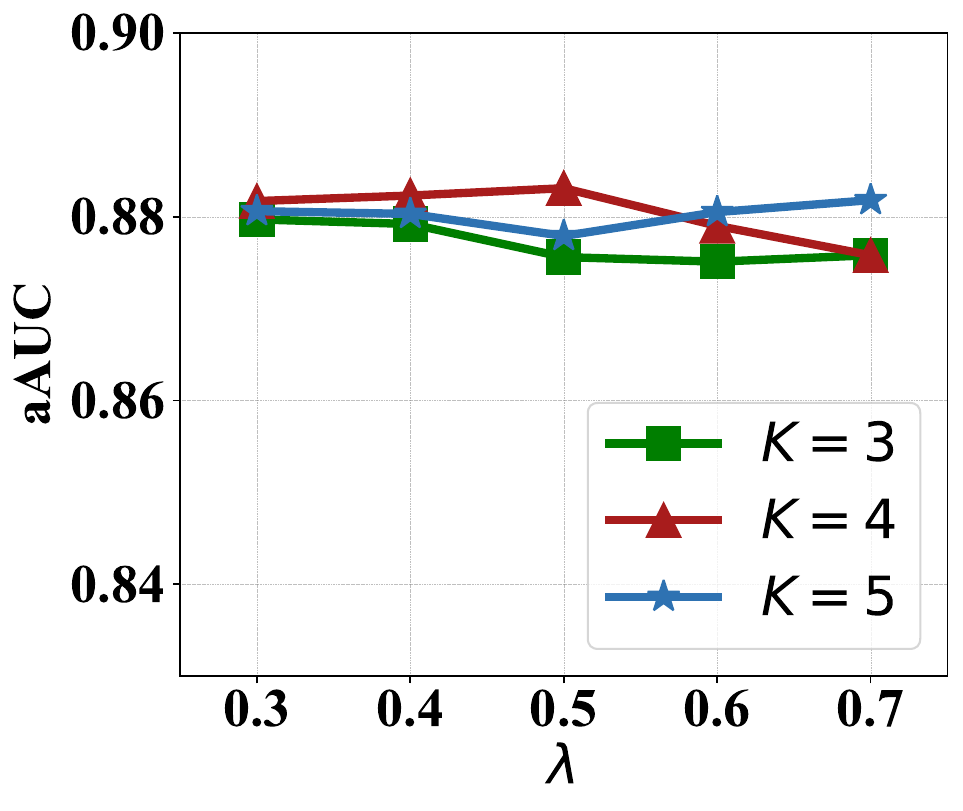}}

    \end{minipage}
    \hfill 
    \begin{minipage}{0.24\textwidth}
        \subfigure[aF1 Score]{\includegraphics[width=\textwidth]{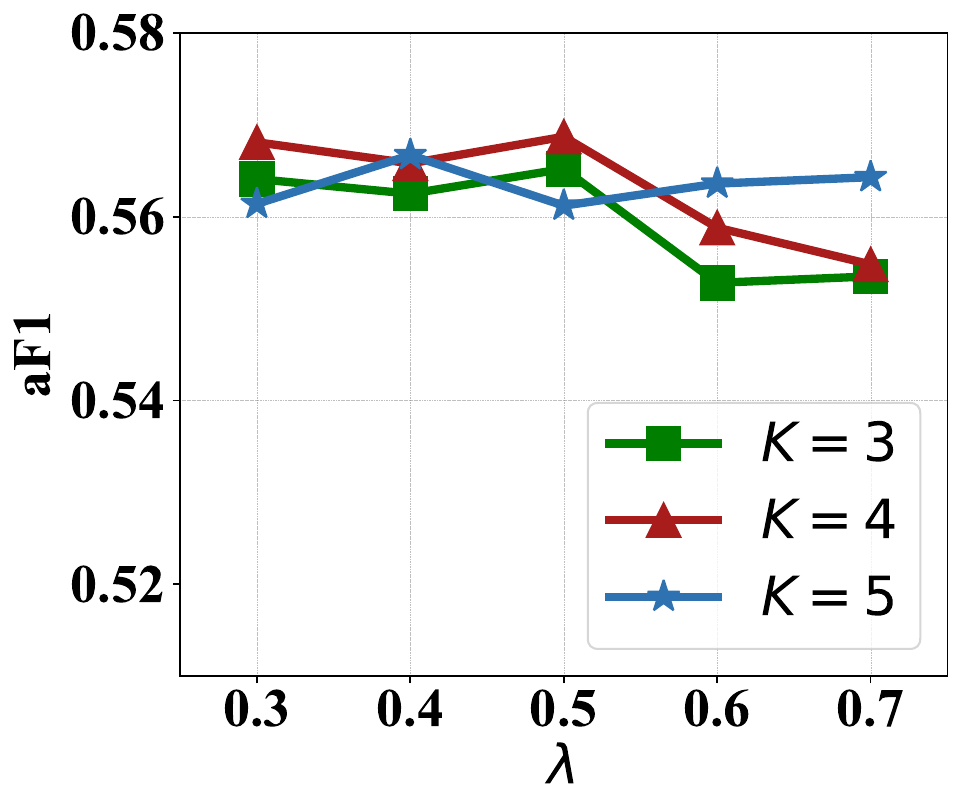}}

    \end{minipage}
    \hfill 
    \begin{minipage}{0.24\textwidth}
        \subfigure[aACC Score]{\includegraphics[width=\textwidth]{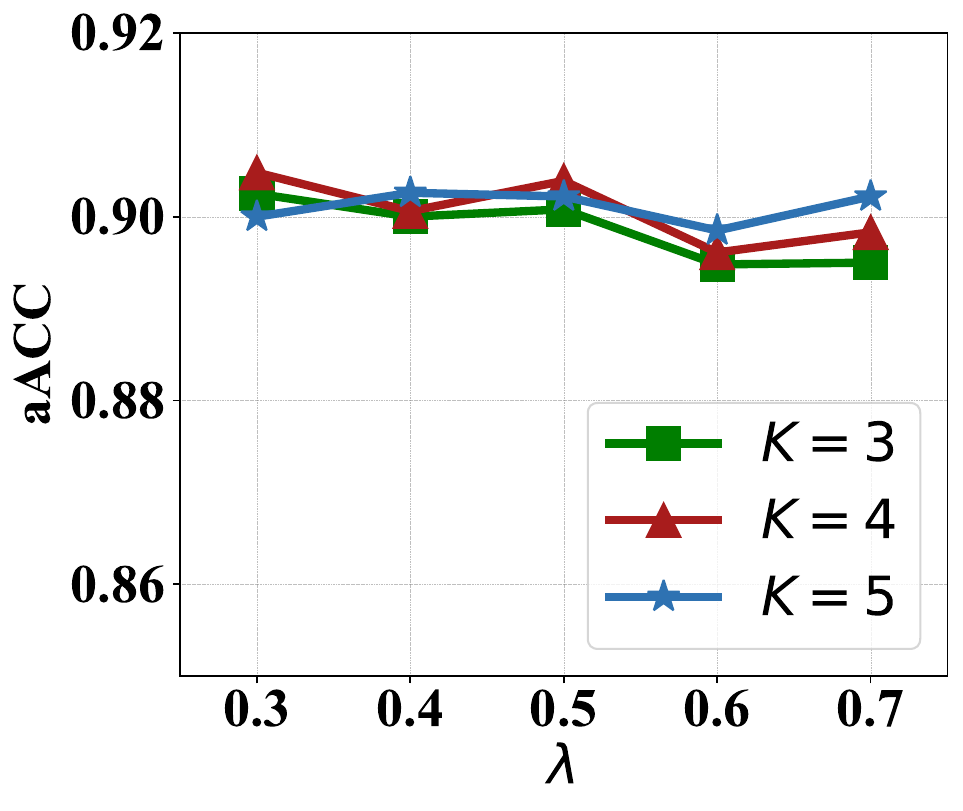}}

    \end{minipage}
    \hfill 
    \begin{minipage}{0.24\textwidth}
        \subfigure[mAP Score]{\includegraphics[width=\textwidth]{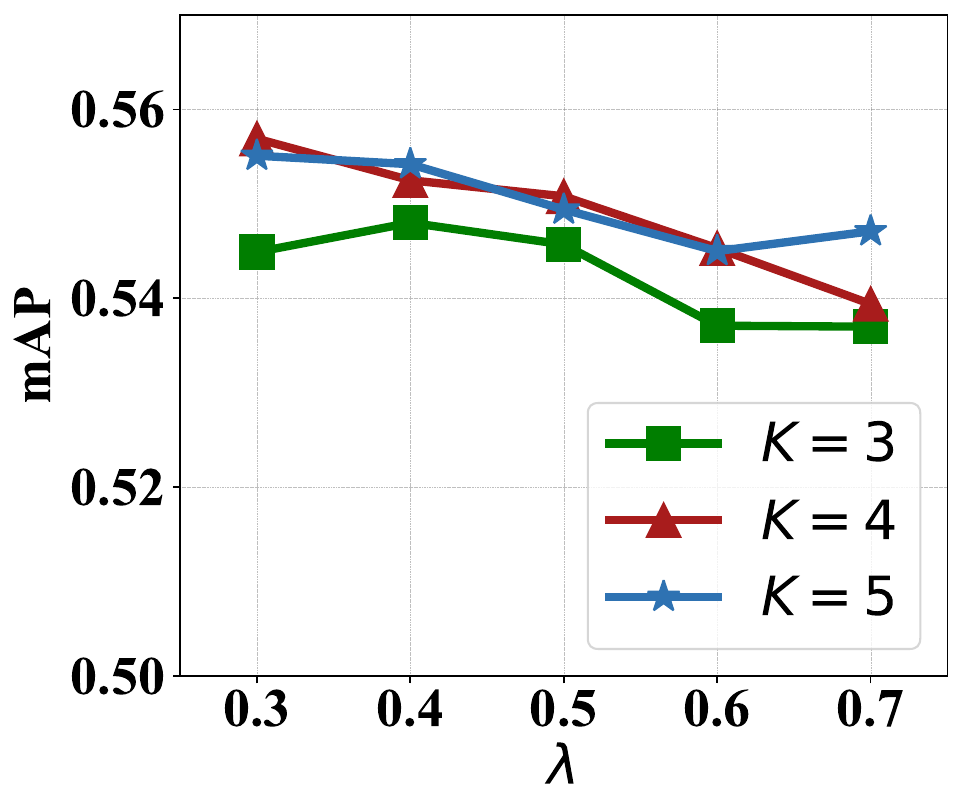}}

    \end{minipage}
    
    \caption{Ablation study on different hyper-parameter combinations, where $\lambda$ is the weight of the first query network during the ``Divide'' stage, and $K$ is the total number of query networks.}

    \label{ablation_hyper}
\end{figure*}

\begin{table}[!t]
\footnotesize
\centering
\caption{Ablation study on the role of the report. Report in the first row means adopting report entities as the input of the text encoder if the sample is with report. ``Label Concat.'' denotes we concatenate corresponding positive labels as the input. ``Partial'' means adopting ``Label Concat.'' only for samples without reports while ``All'' means adopting ``Label Concat.'' regardless of whether accompanied by reports or not.}

\begin{tabular}{ccc|cccc}
\toprule[1.5pt]
\multirow{2}{*}{Report} & \multicolumn{2}{c|}{Label Concat.} & \multirow{2}{*}{aAUC} & \multirow{2}{*}{aF1} & \multirow{2}{*}{aACC} & \multirow{2}{*}{mAP} \\ \cline{2-3}
& Partial & All & & & & \\
\midrule
 & & & 0.8716 & 0.5450 & 0.8939 & 0.5175 \\
\checkmark & & & 0.8742 & 0.5488 & 0.8926 & 0.5217 \\
 & & \checkmark & 0.8733 & 0.5624 & 0.8995 & 0.5458 \\

\midrule
\checkmark & \checkmark & & \textbf{0.8831} & \textbf{0.5687} & \textbf{0.9039} & \textbf{0.5508} \\
\bottomrule[1.5pt]
\end{tabular}
\label{ablation_report}
\end{table}

\subsection{The Robustness under Different Hyper-parameters}
The main hyper-parameters in our UniChest framework are the weight $\lambda$ of the first query network during the ``Divide'' stage, and the total number of query networks $K$. In the previous sections, we set the default values of $\lambda$ as $0.5$ and $K$ as $4$. In Fig.~\ref{ablation_hyper}, we present results for various combinations of $\lambda \in [0.3, 0.7]$ and $K \in [3,5]$, showcasing the average numerical outcomes across six different datasets, including ChestX-ray14, CheXpert, VinDr-CXR, Shenzhen, SIIM-ACR Pneumothorax, and Open-I. For instance, the default setting achieves best numerical results as to aAUC and aF1, while the range of variation for the maximum and minimum values of aAUC and aACC are below 1\%, which validates the robustness of our method under different hyper-parameter combinations.

\subsection{The Role of Report in Pre-training}
As stated in Section~\ref{sec:method}, following the paradigm of vision-language pre-training, we use entities extracted from the report by NER tool as the input of the text encoder $\Phi_{\text{text}}$ if the corresponding report is available, while we concatenate their positive labels as input content for those without reports. In our multi-source pre-training dataset Multi-CXR, there are a total of 685,951 samples, of which 348,900 samples from MIMIC-CXR have reports, meaning that half of the samples come with reports while the rest samples have no corresponding reports. 
It's worth noting that numerous datasets, which currently present only labels and images, such as CheXpert and ChestX-ray14, were originally accompanied by reports. The labels from these datasets were derived using NLP tools akin to our entity extraction method. Nevertheless, due to certain practical restrictions, including privacy concerns, these reports are not publicly available. Consequently, opting for label concatenation as a substitute is a reasonable manner.
To further study the role of the report, we consider three different processing methods. First, we discard the text input directly. Second, we utilize only reports and cancel the input of positive labels for samples without reports. Third, we adopt positive label concatenation as the input for the text encoder for all samples whether they have a report or not. The overall result shown in Table~\ref{ablation_report} drops a little when we underutilize the text input, emphasizing the importance of semantically rich reports in the development of the CXR diagnosis model.

\subsection{Performance under Different Backbones}
We also explore the influence of modality backbones in the whole architecture. Firstly, we substitute ResNet-50 with DenseNet-121~\cite{huang2017densely} for the visual backbone and then replace the default fine-tuned PubMedBERT with ClinicalBERT~\cite{alsentzer-etal-2019-publicly} as textual encoder. As shown in Table~\ref{ablation_encoder}, we notice the average numerical outcomes across six different datasets of the three settings are comparable, demonstrating the insensitivity of backbone selection.

\section{Discussion and Conclusion}
In this paper, we propose a novel Conquer-and-Divide pre-training framework for multi-source Chest X-Rays, namely UniChest, which is among the first attempts to fuse and utilize CXR samples from various origins harmonically. 
Our method effectively balances the benefits of multi-source data while minimizing the negative impacts the inter-source heterogeneity brings. In the ``Conquer'' stage, UniChest enhances feature extraction by sharing model components across different sources. In the ``Divide'' stage, it employs a mixture of deep query modules and utilizes a novel source-contrastive learning loss to isolate source-specific patterns, further reducing cross-source interference.
Through extensive experiments on a range of benchmark datasets, we show the robustness and effectiveness of UniChest under diverse settings in CXR diagnosis.

\begin{table}[!t]
\footnotesize
\centering
\caption{Ablation study on different modality encoders.}
\setlength{\tabcolsep}{9pt}
\begin{tabular}{c|cccc}
\toprule[1.5pt]
Method & aAUC & aF1 & aACC & mAP \\ 
\midrule
w/ DenseNet-121 & 0.8794 & 0.5531 & \textbf{0.9078} & 0.5410 \\
w/ ClinicalBERT &  \textbf{0.8849} & 0.5643  & 0.8997 & 0.5427 \\
\midrule
\textbf{UniChest} & 0.8831 & \textbf{0.5687} & 0.9039 & \textbf{0.5508} \\
\bottomrule[1.5pt]
\end{tabular}
\label{ablation_encoder}
\end{table}

Despite its powerful performance, UniChest still has some limitations in its design and application. Firstly, similar to MedKLIP and KAD, UniChest is limited to generating coarse-grained grounded heatmaps by utilizing cross-attention maps, which falls short of meeting the requirements for precise pixel-level segmentation. Therefore, the development of a comprehensive universal CXR model that combines both classification and fine-grained lesion grounding is a promising avenue for benefiting the medical community, which would not only offer efficiency but also ensure trustworthiness in daily applications. Secondly, although the Conquer-and-Divide framework is precise and meaningful and we have proposed one implementation paradigm within this framework. There is room for further exploration of other concrete frameworks that align with this spirit. We hope that our UniChest will inspire the exploration and utilization of multi-source CXRs.

\appendices

{
\bibliographystyle{IEEEtran}
\bibliography{reference.bib}
}

\end{document}